\documentclass[sigconf,screen]{acmart}
\AtBeginDocument{%
  }
\usepackage{graphicx} 

\usepackage{balance}
\usepackage{multicol} 

\usepackage{lettrine}

\usepackage{multirow, soul}
\usepackage{booktabs}
\usepackage{xcolor}
\usepackage{array}
\usepackage{amsmath}
\usepackage{graphicx}
\usepackage{colortbl}
\usepackage{hhline}
\setcopyright{rightsretained}
\usepackage{etoolbox}
\usepackage{ccicons}  
\usepackage{etoolbox} 

\usepackage{csquotes}
\usepackage{multicol}

\copyrightyear{2025}
\acmYear{2025}
\setcopyright{rightsretained}

\acmConference[WWW Companion '25]{Companion Proceedings of the ACM Web Conference 2025}{April 28-May 2, 2025}{Sydney, NSW, Australia}
\acmBooktitle{Companion Proceedings of the ACM Web Conference 2025 (WWW Companion '25), April 28-May 2, 2025, Sydney, NSW, Australia}
\acmDOI{10.1145/3701716.3715526}
\acmISBN{979-8-4007-1331-6/2025/04}

\makeatletter
\gdef\@copyrightpermission{
  \begin{minipage}{0.2\columnwidth}
   \href{https://creativecommons.org/licenses/by-nc-sa/4.0/}{\includegraphics[width=0.90\textwidth]{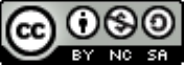}}
  \end{minipage}\hfill
  \begin{minipage}{0.8\columnwidth}
   \href{https://creativecommons.org/licenses/by-nc-sa/4.0/}{This work is licensed under a Creative Commons Attribution-NonCommercial-ShareAlike International 4.0 License.}
  \end{minipage}
  \vspace{5pt}
}
\makeatother

\begin{document}



\title[Towards Safe Synthetic Image Generation On the Web]{Towards Safe Synthetic Image Generation On the Web: A
Multimodal Robust NSFW Defense and Million Scale Dataset}
\author{Muhammad Shahid Muneer}
\orcid{0009-0007-6093-8251}
\affiliation{%
  \institution{DASH Lab \\ Computer Science and Engineering Department \\ Sungkyunkwan University}
  \city{Suwon}
  \country{Republic of Korea}
}
\email{shahidmuneer@g.skku.edu}

\author{Simon S. Woo}
\orcid{0000-0002-8983-1542}
\authornote{Corresponding author. Email: swoo@g.skku.edu (Simon S. Woo)}
\affiliation{%
  \institution{DASH Lab \\ Computer Science and Engineering Department \\ Sungkyunkwan University}
  \city{Suwon}
  \country{South Korea}
}
\email{swoo@g.skku.edu}


\begin{abstract}

In the past years, we have witnessed the remarkable success of Text-to-Image (T2I) models and their widespread use on the web. Extensive research in making T2I models produce hyper-realistic images has led to new concerns, such as generating Not-Safe-For-Work (NSFW) web content and polluting the web society. To help prevent misuse of T2I models and create a safer web environment for users features like NSFW filters and post-hoc security checks are used in these models. However, recent work unveiled how these methods can easily fail to prevent misuse. In particular, adversarial attacks on text and image modalities can easily outplay defensive measures. 
Moreover, there is currently no robust multimodal NSFW dataset that includes both prompt and image pairs and adversarial examples. This work proposes a million-scale prompt and image dataset generated using open-source diffusion models. Second, we develop a multimodal defense to distinguish safe and NSFW text and images, which is robust against adversarial attacks and directly alleviates current challenges. Our extensive experiments show that our model performs well against existing SOTA NSFW detection methods in terms of accuracy and recall, drastically reducing the Attack Success Rate (ASR) in multimodal adversarial attack scenarios. Code:
~\href{https://github.com/shahidmuneer/multimodal-nsfw-defense.}{\textbf{GitHub}}.

\end{abstract}




\maketitle
\textbf{\textcolor{red}{Disclaimer:}} \textcolor{red}{This paper contains explicit NSFW materials that can be disturbing or offensive to some readers. We include them solely for research purposes and to develop multimodal defenses.}


\begin{figure}[h]
  \centering
  \includegraphics[width=1\linewidth]{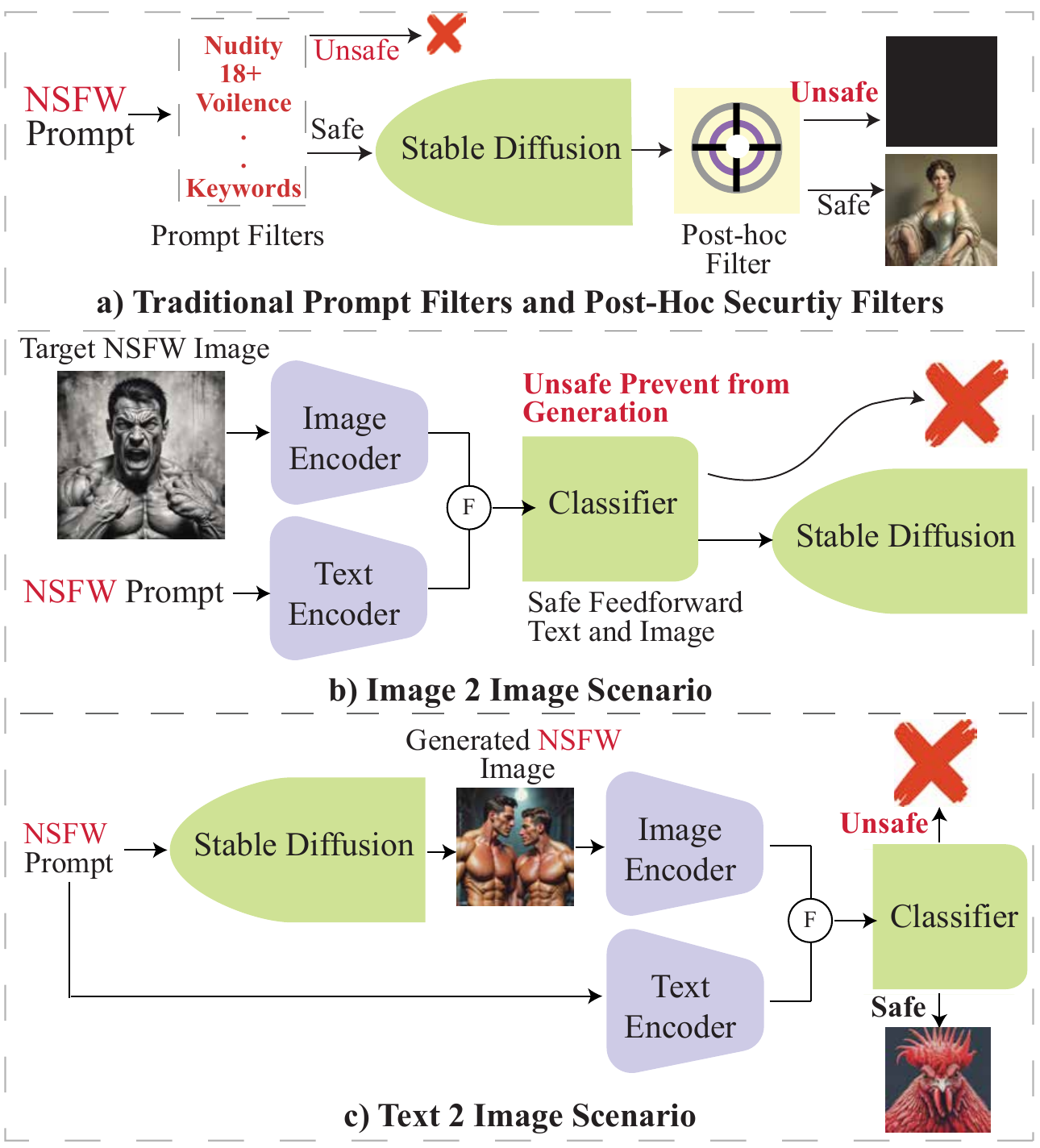}
  \caption{a) Existing prompt filters and post-hoc security filters are employed in AI Image generation models, while b) and c) are our proposed implementations for context-dependent multimodal defense in T2I and I2I models respectively.}

  \Description{}
\label{fig:proposed_model}
\end{figure}
\section{Introduction}
AI image generation advancements started with Generative Adversarial Networks (GANs)~\cite{goodfellow2014generativeadversarialnetworks}. Currently, the new generative AI models such as Stable Diffusion~\cite{rombach2021highresolution} can produce increasingly realistic AI content. Currently, diffusion-based models play an important role on the web by rapidly producing highly realistic online content, and they are being shared through various social networking platforms. In particular, the introduction of Transformer network~\cite{vaswani2023attentionneed}, CLIP-ViT~\cite{pmlr-v139-radford21a} and multimodal approaches~\cite{turk2014multimodal,rombach2021highresolution,ngiam2011multimodal} have paved the way for human-interactive Text-to-Image (T2I) and Image-to-Image (I2I) models. T2I models can generate images in a closer context of user-provided personalized text prompts. And, this enables users to create and share diverse online content using T2I and I2I on the web. The applicability of T2I and I2I models in real life gave companies such as Midjourney~\cite{midjourney} a wide path to deliver high-quality T2I and I2I models. The images generated using open-source models such as Stable Diffusion and commercial T2I models have a variety of benefits.

However, positive use can come with a negative aspect; users with malicious intentions can use T2I or I2I models to generate close-to-real NSFW images using the NSFW prompts, where NSFW prompts or images contain nudity, pornography, violence, ethnicity, racism, and offensive words or images that are not desirable for the safe online scenarios. Such content has skyrocketed and started polluting and jeopardizing the health of the online community. 

Although both open-source and commercial AI image generation methods have employed prompt filters and post-hoc security filters to prevent the generation of unsafe NSFW contents, such methods are shown to be defenseless against multimodal adversarial attacks~\cite{yang2024mmadiffusion}. Similarly, post-hoc security filters detect generated images that contain NSFW content; however, these are vulnerable to adversarial attacks~\cite{NEURIPS2023_dd83eada} and do not work well in classifying context and text-to-image correlation. The multimodal adversarial attack~\cite{yang2024mmadiffusion} on both text and image modalities seeks a multimodal solution.
These limitations motivate our investigation into a reliable multimodal defense against multimodal adversarial attacks and NSFW content. The intuition behind the proposed multimodal defense framework is not only to protect AI image generation models from multimodal adversarial attacks but also to propose a defense mechanism that learns the relationship and context of both text and image modalities in filtering NSFW content instead of using separate prompt filters and post-hoc security filters. Specifically, in this research, we investigate and curate a multimodal dataset that contains NSFW prompts, corresponding synthetic images, and adversarial text and images. 

The proposed approach is particularly suitable for distinguishing between NSFW text and image from safe content based on the context and correlation because of our large-scale prompt image paired NSFW dataset with adversarial examples and multimodal cross-attention strategy. We 
empirically demonstrated the ability of the proposed model to prevent adversarial NSFW text and images along with natural profanities. As shown in Table~\ref{tab:table3} Attack Success Rate (ASR) on MMA-Attack~\cite{yang2024mmadiffusion} has exceptionally reduced. Our main contributions are summarized as follows.

\begin{itemize}
    \item We provide a large-scale multimodal text and image profanity dataset NSFWCorpus, consisting of 1 million NSFW text and image pairs, which not only contains real samples generated using diffusion models but also NSFW images containing both adversarial and normal profanity.  
    \item We propose a multimodal defense mechanism for T2I and I2I models to guard them against profanity and adversarial attacks effectively. The proposed model can be employed with any open-source or commercial web applications.
    \item Extensive experiments demonstrate the effectiveness of our multimodal approach against SOTA adversarial attacks and NSFW content generation. Particularly, our proposed approach performed strongly in preventing multimodal adversarial attacks.
\end{itemize}

\section{Method}
The proposed approach uses prompt-to-image correlation to clear context for classifying NSFW from safe images. The training data is collected and synthesized, and it contains over 1 million text and image pairs shown in Table~\ref{tab:table1}. The dataset generation pipeline is described in Section 2.2, and the proposed multimodal defense is discussed in Section 2.3.

\begin{figure}[h]
\centering
\includegraphics[width=1\linewidth]{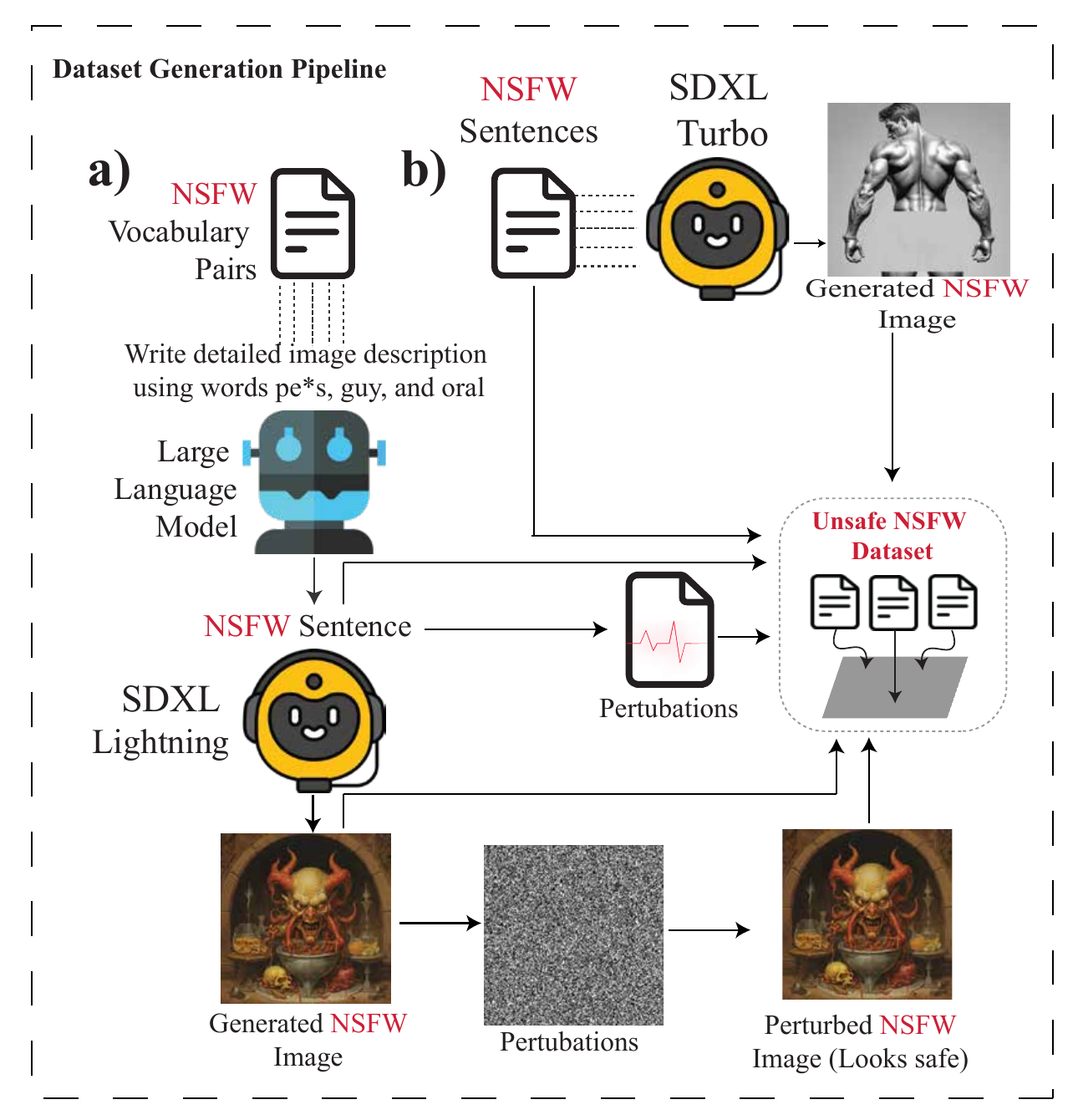}
\caption{Overall dataset generation pipeline. }
\label{fig:nsfw_dataset_pipeline}
\end{figure}

\subsection{Dataset Generation Pipeline}

Our dataset generation pipeline consists of two methods to generate sensitive content containing NSFW text and images. In the first method shown in Fig.~\ref{fig:nsfw_dataset_pipeline}, we utilized NSFW vocabulary~\cite{surgeai_profanity} ${V_\mathrm{NSFW}}$ which contains 1,599 pairs of words consisting of sexual, bodily fluids, sexual orientation, mental disability, physical disability, racial-ethnic, racism, physical disability, physical attributes, animal references, religious offenses, and political related NSFW words. We utilized it ${V_\mathrm{NSFW}}$ to form a sentence ${S_\mathrm{NSFW}}$ defined in Eq.~\ref{eq:nsfw_s} below: 
{
\begin{equation}
\label{eq:nsfw_s}
S_\mathrm{NSFW} = \sum_{i=1}^{N} w_i \cdot v_i, \quad v_i \in V_{NSFW}, \quad w_i \in \{0, 1\},
\end{equation}
}
\noindent where \( v_i \) represents the \( i \)-th word from the NSFW vocabulary $V_\mathrm{NSFW}$, and \( w_i \) is a binary indicator, where \( w_i = 1 \) if the word \( v_i \) is selected for the sentence, and \( w_i = 0 \) otherwise. \( N \) is the total number of words in the vocabulary \( V_\mathrm{NSFW} \).

We feed forward the selected sentence to an open-source Large Language Model (LLM) $LLM_\mathrm{misteral}$ Mistral-7B-Instruct-v0.2 as shown in Eq.~\ref{eq:nsfw_llm}:
{\begin{equation}
\label{eq:nsfw_llm}
 P_\mathrm{NSFW} = LLM_\mathrm{mistral}(concat(\text{``Description of image'' },  S_\mathrm{NSFW}))
\end{equation}
}
\noindent where \( LLM_\mathrm{mistral} \) represents LLM and ${P_\mathrm{NSFW}}$ represents prompt containing NSFW image description. The generated prompt is a meaningful description of the image that contains the context and detailed explanation of the image. ${P_\mathrm{NSFW}}$ is then forwarded to an open-source T2I model SDXL Lightning for high-quality and detailed image generation by disabling the prompt and post-hoc security filters. The image generation process is defined in the following Eq.~\ref{eq:nsfw_image_generation}:
{
\begin{equation}
\label{eq:nsfw_image_generation}
I_{NSFW} = SDXL_\mathrm{lightning}(P_\mathrm{NSFW}),
\end{equation}
}
\noindent where ${I_\mathrm{NSFW}}$ represents the image generated using the diffusion model. We performed adversarial attacks on both target prompt ${P_\mathrm{NSFW}}$ and generated Image ${I_\mathrm{NSFW}}$ resulting in an adversarial prompt  ${P_\mathrm{adv_\mathrm{NSFW}}}$ and adversarial image ${I_\mathrm{adv_\mathrm{NSFW}}}$ as shown in Fig.~\ref{fig:nsfw_dataset_pipeline} and Eq.~\ref{eq:adv_prompt}:

\begin{table}[b!]
\centering
\caption{Details of Proposed  and Collected Dataset}
\begin{tabular}{lrr}
\toprule
\textbf{Type}                      & \textbf{\# Images}    & \textbf{\# Prompts}   \\ 
\midrule
- Synthetic Diffusion Images~\cite{wangDiffusionDBLargescalePrompt2022} & 400,000 & 400,000 \\ 
- Microsoft COCO~\cite{lin2015microsoftcococommonobjects}               & 100,000 & 100,000 \\  
\textbf{Total Real}                & \textbf{500,000}      & \textbf{500,000}      \\ 
- Generated (SDXL Lightning)         & 180,000                & 180,000                \\ 
- Generated (SDXL Turbo)             & 300,000               & 300,000               \\ 
- Perturbed (Image/Text Only)        & 20,000                & 20,000                \\ 
\textbf{Total NSFW}                & \textbf{500,000}      & \textbf{500,000}      \\ 
\bottomrule
\textbf{Grand Total}               & \textbf{1,000,000}      & \textbf{1,000,000}      \\ 
\bottomrule
\end{tabular}
\label{tab:table1}
\end{table}

{
\begin{equation}
\label{eq:adv_prompt}
P_\mathrm{adv_\mathrm{NSFW}} = P_\mathrm{random} + \delta_P, \; P_\mathrm{random} \sim \mathcal{U}(P_\mathrm{NSFW}), \; \|\delta_P\| \leq \epsilon_P,
\end{equation}
}

\noindent where \( P_\mathrm{random} \) is the randomly chosen prompt from database \( P_\mathrm{NSFW}\) using function \( \mathcal{U} \), and \( \delta_P \) is perturbation added to the original prompt, designed to be imperceptible or semantically subtle. And, \( \epsilon_P \) is the maximum allowed perturbation norm to ensure imperceptibility, and \( \|\delta_P\| \) is the norm of the perturbation, which must be less than or equal to \( \epsilon_P \).

Next, the adversarial prompt \( P_\mathrm{adv_\mathrm{NSFW}} \) is crafted to mislead the diffusion model \( SDXL_\mathrm{lightning} \) into generating a manipulated or altered image as follows:
{
\begin{equation}
\label{eq:adv_image}
I_\mathrm{adv, NSFW} = I_\mathrm{random} + \delta_I, \quad I_\mathrm{random} \sim \mathcal{U}(I_\mathrm{NSFW}), \quad \|\delta_I\| \leq \epsilon_I
\end{equation}
}

\noindent where \( I_\mathrm{adv_\mathrm{NSFW}}\) is the perturbed image, \( \delta_I \) are perturbations added to the generated image \( I_\mathrm{NSFW}\), designed to be imperceptible while altering the classification or intended perception of the image. Also, \( \epsilon_I \) is the maximum allowed perturbation norm for the image perturbation, and \( \|\delta_I\| \) norm of the perturbation, which must be less than or equal to \( \epsilon_I \). Both \( P_\mathrm{random} \) and \( I_\mathrm{random} \) are chosen randomly from pool of \( P_\mathrm{NSFW} \) and \( I_\mathrm{NSFW} \). We have used MMA-Diffusion~\cite{yang2024mmadiffusion} attacks to generate NSFW prompts and images.

While process b) shown in Fig.~\ref{fig:nsfw_dataset_pipeline} uses NSFW sentences scrapped online ${P_\mathrm{srp_\mathrm{NSFW}}}$ to generate images using SDXL Turbo ${I_\mathrm{srp_\mathrm{NSFW}}}$, and ${P_\mathrm{safe}, I_\mathrm{safe}}$ represent safe to use prompts and images outsourced from open-source datasets. Hence, the dataset can be defined as follows in Eq.~\ref{eq:dataset_definition}:
{\begin{align}
\label{eq:dataset_definition}
\mathcal{D}_\mathrm{NSFW} = & \{(P_\mathrm{safe}, I_\mathrm{safe}), (P_\mathrm{NSFW}, I_\mathrm{NSFW}), \notag \\
& (P_\mathrm{adv_\mathrm{NSFW}}, I_\mathrm{adv_\mathrm{NSFW}}), (P_\mathrm{srp_\mathrm{NSFW}}, I_\mathrm{srp_\mathrm{NSFW}})\}
\end{align}}

\begin{figure}[htb]
  \centering
  \includegraphics[width=1\linewidth]{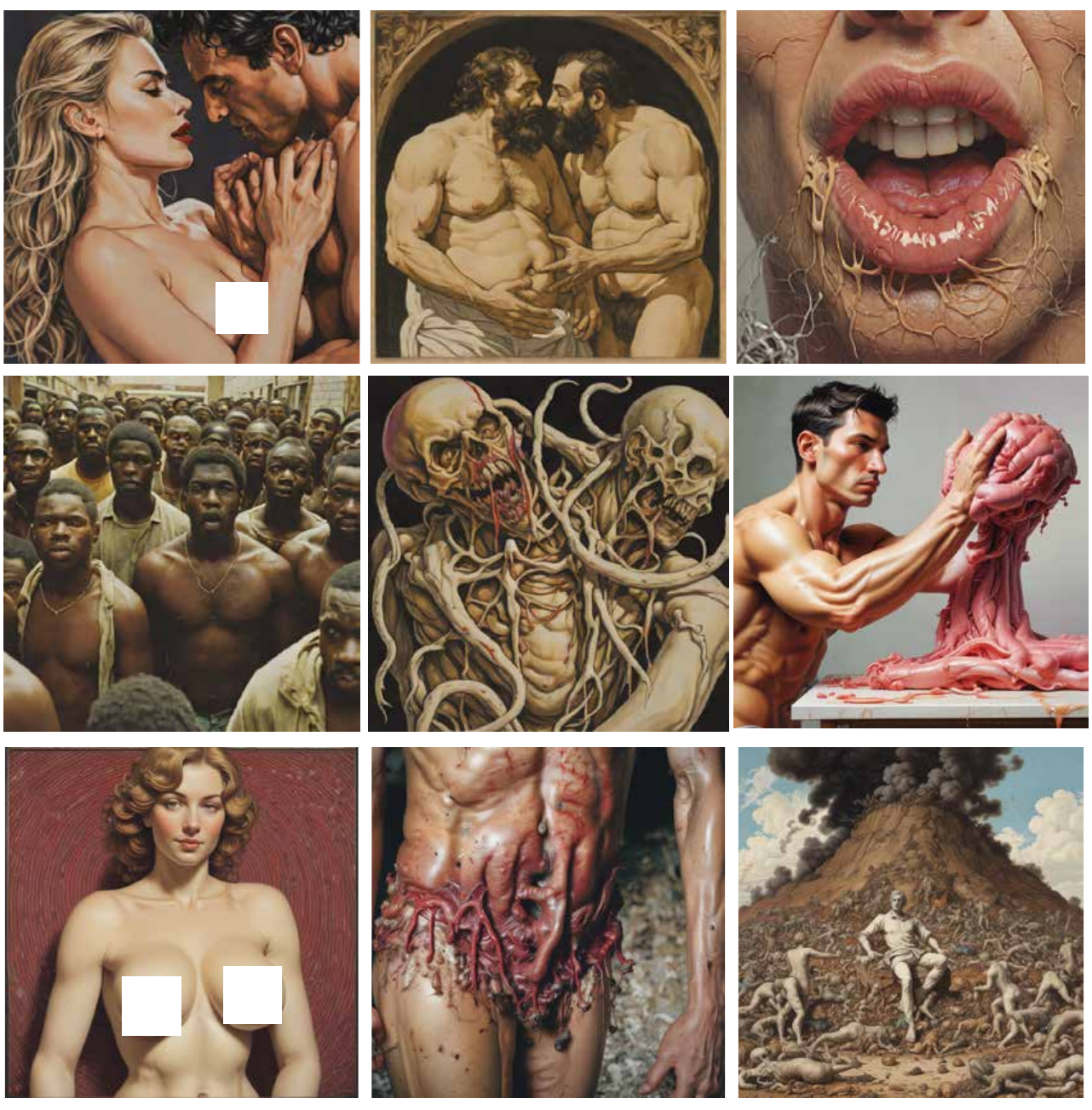}
  \caption{Illustration of images depicting racism, nudity, violence, and gender discrimination context included in the generated dataset. }
  \Description{Illustration of images depicting racism, nudity, violence, and gender discrimination context included in the generated dataset.}
  \label{fig:dataset_samples}
\end{figure}

Fig.~\ref{fig:dataset_samples} shows some samples from the generated dataset, consisting of racism, gender discrimination, nudity, and violence, generated using SDXL Turbo and SDXL lightning. 

\subsection{Proposed Multimodal Model}
The proposed multimodal defense model consists of a text and image encoder, as shown in Fig.~\ref{fig:proposed_model}. We utilized multi-head attention for multimodal fusion and then forwarded the CLS tokens for final classification. To explain this further, let us consider the input to the model to be either an original or adversarial prompt (\(P_\mathrm{NSFW}\), (\(P_\mathrm{safe}\), and \(P_\mathrm{adv}\)). We utilized CLIP-ViT/32 Large for extracting text and image features. For text feature extraction, we first converted the prompts into tokens and then forwarded them to the feature extractor as shown in Eq.~\ref{eq:eq_10} below: 
{\begin{align}
\label{eq:eq_10}
F_t = F_{\text{text}}(\text{Tokenizer}(P_{\mathrm{NSFW}_i}, P_\mathrm{safe_i},P_\mathrm{adv_i})), \quad F_t \in \mathbb{R}^d
\end{align}

}
\begin{figure}[h!]
  \centering
  \includegraphics[width=1\linewidth]{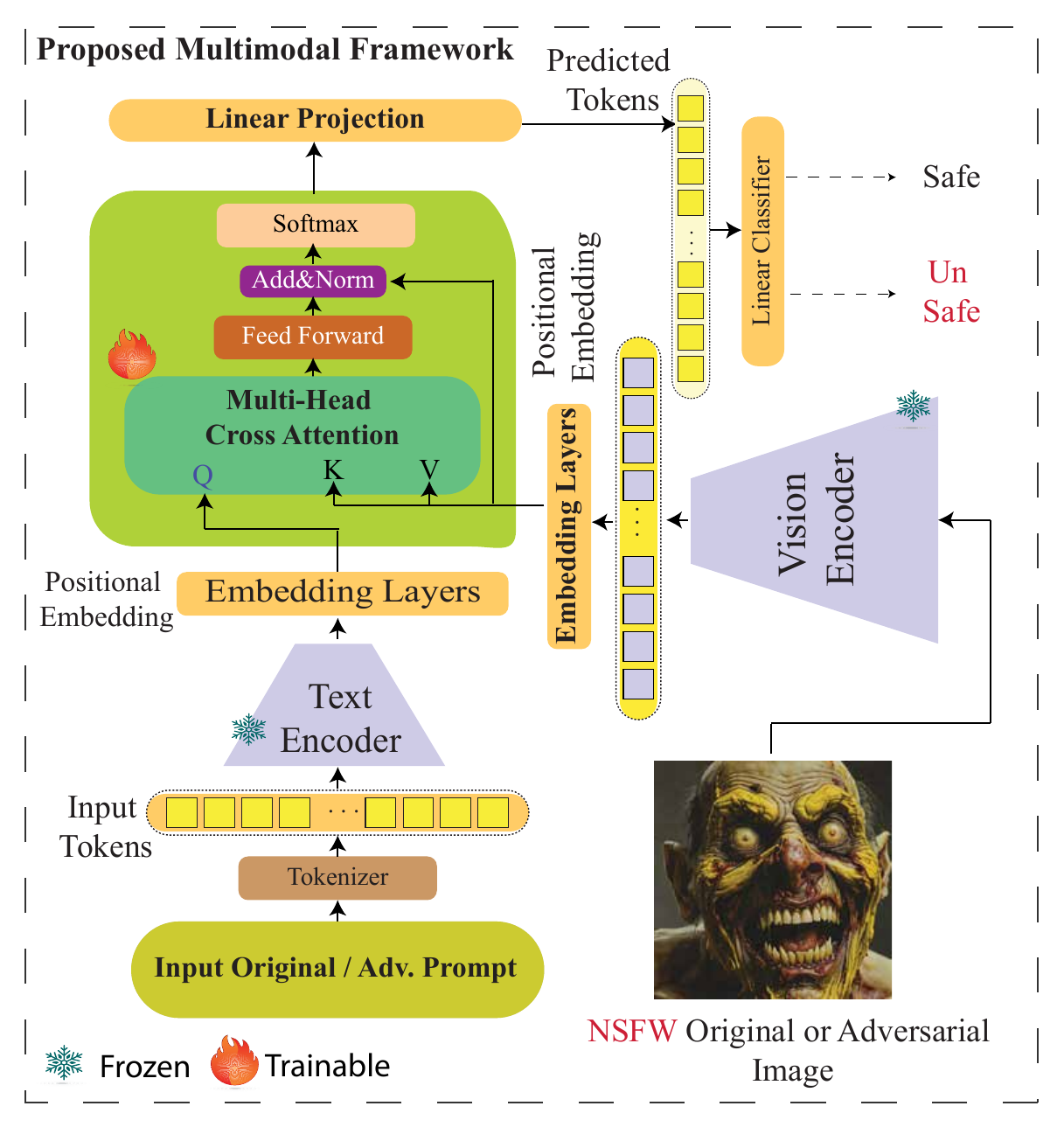}
  \caption{Illustration of a high-level diagram of our proposed framework to classify safe and NSFW content. }
  \Description{Low-Level diagram of the proposed framework to classify safe and NSFW content using text and vision encoders with cross-attention mechanism.}
  \label{fig:proposed_model}
\end{figure}

\noindent where \(d\) is the embedding dimension and ${F_\mathrm{text}}$ is the text encoder. 

The embeddings \(E_t\) are propagated through linear layers to obtain positional embeddings as follows:
\begin{align}
\label{eq:eq_8}
H_t = f_\mathrm{\text{embed}}(F_t), \quad H_t \in \mathbb{R}^{L \times d}
\end{align}

Similarly as shown above, we forwarded  \(I_\mathrm{NSFW}\), \(I_\mathrm{safe}\), and \(I_\mathrm{adv}\) for feature extraction in vision encoder.

\begin{table*}[htbp]

\centering
\caption{Comparison of unimodal and multimodal attacks listed on the leftmost column with subsequent text, and multimodal NSFW content filtration methods on the top row based upon Attack Success Rate (ASR). ASR-1 is attack success rate w.r.t. 1 prompt while ASR-4 is w.r.t. 4 prompts in a single attack. Bold represents the best, \underline{underlined} indicates the second best, and $\downarrow$ is a decrease in ASR. }
\label{tab:table2}
\begin{tabular}{cc|cccccc|cc}
\toprule
\multicolumn{1}{c}{\textbf{}} & \multicolumn{1}{c|}{\textbf{}} & \multicolumn{8}{c}{\textbf{Defense Method}} \\
\multicolumn{1}{c}{} & \multicolumn{1}{c}{} & \multicolumn{2}{c}{Q16 (Text)~\cite{q16} } & \multicolumn{2}{c}{MHSC (Text)~\cite{QSHBZZ23} } & \multicolumn{2}{c}{SC (Text)~\cite{CompVisStableDiffusionSafetyChecker}} & \multicolumn{2}{c}{\textbf{Ours  (Text\&Image) $\downarrow$ }} \\
\cmidrule(lr){3-4} \cmidrule(lr){5-6} \cmidrule(lr){7-8} \cmidrule(lr){9-10}
\textbf{} & \textbf{Attack Model} & ASR-4 & ASR-1 & ASR-4 & ASR-1 & ASR-4 & ASR-1 & ASR-4 & ASR-1 \\
\midrule
\textcolor{blue}{CVPR'23}  & I2P (Text)   ~\cite{QSHBZZ23}   & 9.60 & 8.24 & \underline{5.97} & 4.48 & 6.31 & \underline{3.30} & \textbf{3.38} & \textbf{2.34} 
  \\
\textcolor{blue}{NIPS'23}  & Greedy (Text)~\cite{Zhuang_2023_CVPR} & 3.20 & 1.15 &  \underline{1.88} & \underline{0.67} & 1.92 & 0.70 & \textbf{0.21}   & \textbf{1.13} \\
\textcolor{blue}{NIPS'23}  & Genetic (Text)~\cite{Zhuang_2023_CVPR}& \underline{1.57} & \underline{0.57} & 3.44 & 1.26 & 2.08 & 0.75 & \textbf{0.11}  & \textbf{0.15}  \\
\textcolor{blue}{NIPS'23} & QF-PGD (Text)~\cite{Zhuang_2023_CVPR} & 2.24 & 0.78 & \textbf{1.54} & \underline{0.46} & 1.63 & 0.51 & \underline{2.29} & \textbf{0.31}  \\
\textcolor{blue}{CVPR'24}  & MMA-DIFFUSION (Text \& Image)~\cite{yang2024mmadiffusion} & 84.90 & 73.23 & 84.80 & 75.10 & \underline{80.40} & \underline{54.20} & \textbf{3.78}   & \textbf{2.22 } \\
\bottomrule

\end{tabular}

\label{tab:table3}
\end{table*}

The embeddings \(E_t\) are propagated through linear layers to obtain positional embeddings:
\begin{align}
\label{eq:eq_8}
H_i = f_\mathrm{\text{embed}}(Fi), \quad H_i \in \mathbb{R}^{L \times d}
\end{align}

The obtained text embeddings \(H_t\) and vision embeddings \(H_i\) interact via a multi-head cross-attention mechanism to learn multimodal representations. While Queries (\(Q\)) are derived from \(H_t\), Keys (\(K\)) and Values (\(V\)) are derived from \(H_i\) shown below in Eq.~\ref{eq:eq_Q}:
{\begin{align}
\label{eq:eq_Q}
Q = W_Q H_t, \quad K = W_K H_i, \quad V = W_V H_i,
\end{align}
\noindent where \(W_Q, W_K, W_V \in \mathbb{R}^{d \times d}\) are learnable parameters. An attention score is computed as shown in Eq.~\ref{eq:eq_9}:
\begin{align}
\label{eq:eq_9}
A = \text{Softmax}\left(\frac{QK^T}{\sqrt{d}}\right), \quad A \in \mathbb{R}^{L \times N}
\end{align}}

Finally, with added normalization to the cross-attention output \( O \), we forward it to a residual connection of \( H_t \) and normalize it to feed forward to the Linear layer for classification.

\section{Experiments}

\subsection{Experimental Settings}
We proposed text and image pair dataset in Table~\ref{tab:table1} for training, and split the dataset into 70\% for training and 30\% for validation.  For evaluation and baseline comparison on the test set, we used a mix of Human prompts (I2P)~\cite{QSHBZZ23}, MMA-Attack~\cite{yang2024mmadiffusion}, QF-Attack~\cite{Zhuang_2023_CVPR}, and PGD attack ~\cite{Zhuang_2023_CVPR} to compare our multimodal defense with existing unimodal SOTA defenses~\cite{QSHBZZ23,q16} and Stable diffusion inbuild defenses (SC)~\cite{CompVisStableDiffusionSafetyChecker}.  We used Attack Success Rate (ASR) to evaluate the model~\cite{yang2024mmadiffusion}, and Accuracy metrics. 

\subsection{Benchmark Performance }
Table~\ref{tab:table3} compares single modality filters listed on the top row as defense methods, and uni- or multi-modal attacks such as MMA-Diffusion~\cite{yang2024mmadiffusion} are shown in the first column on the left with our multimodal defense on the right column. Our proposed model shows a decline in adversarial attack success rate in both ASR-1 and ASR-4~\cite{yang2024mmadiffusion} cases as compared to traditional prompt filters. We generated images utilizing prompts for multimodal defense evaluation and provided a comparison with unimodal filters. This shows how effectively our proposed adversarial training and multimodal context-aware model can be used to prevent unsafe image generation. The robustness against adversarial attacks and reduction in attack success rate, particularly in multimodal attack scenarios such as MMA-Diffusion, is visible. On the other hand, QF-PGD shows a decrease in ASR-4 metrics, while in all other scenarios, the proposed model shows an exceptional reduction in both ASR-1 and ASR-4. Also, Table~\ref{tab:table2} shows an increase in accuracy, recall, and f1-score metrics comparison with our model trained on the proposed dataset and tested on human-generated prompts~\cite{q16}. 



\begin{table}[h]
\centering
\caption{Performance comparison with SOTA models. Where \textbf{bold} is best and \underline{underline} is the second best performance. }

\renewcommand{\arraystretch}{1.2} 
\setlength{\tabcolsep}{2pt}       

\begin{tabular}{c|cccc}

\toprule
\textbf{Method} & \textbf{Accuracy \%} & \textbf{Precision \%} & \textbf{Recall \%} & \textbf{F1-Score \%}  \\
\midrule
 Safety Filters & 75.57 &  59.65 & 52.25 & 55.05 \\
  Q16 & 70.79 & 49.10 & 73.01 & 59.28  \\
  Fine-tuned Q16 & 88.66 & 77.67 & \underline{83.18} &  80.00 \\
  Multi-head SC &  \underline{90.87} &  \textbf{87.87} & 78.58 & \underline{82.00}  \\
\bottomrule
  \textbf{Ours} & \textbf{92.31} & \underline{85.24} & \textbf{92.89} &  \textbf{88.01}    \\
\bottomrule
\end{tabular}
\label{tab:table2}
\end{table}

\section{Ethical Consideration \& Social Impact}
Our study firmly follows ethical standards~\cite{ethicalResearch} for dataset generation and collection, utilizing only non-identifiable, AI-generated content and publicly sourced prompts, additionally consulting the IRB from our institution. No real individuals were involved, ensuring privacy and responsible adherence to data integrity standards. The generated dataset is available exclusively for research purposes and is not publicly released to ensure ethical usage. Both the dataset and the proposed model contribute to social good by strengthening Stable Diffusion-based models against multimodal adversarial attacks.

\section{Limitation \& Future Work}
While the proposed multimodal defense and the million-scaled dataset shows effective results in preventing multimodal and unimodal attacks, there are some notable limitations. First, the proposed dataset contains the English language, and in real-world scenarios, defensive mechanisms can pose several languages. Second, different cultures have different contexts and niches. Although the proposed dataset can effectively defend against NSFW with diverse context considerations, there is a need for a more diverse dataset. Second, safe examples are taken from the MS Coco dataset, which can be further improved by adding more datasets with alignment to unsafe contexts. 

In future work, we plan to extend our dataset to include languages and their corresponding images in context. We also intended to investigate regional and cultural biases to ensure broader applications of the proposed model and the dataset. Another key direction we will consider is to utilize Large Language Models (LLMs) for effective and improved content moderation with Chain of Thoughts reasoning. Finally, we intend to extend our proposed model to video, audio, and other misinformation remains an important challenge to follow for building a safer web.

\section{Conclusion}
We proposed a large-scale context-oriented prompt and image-paired dataset for NSFW content moderation. While existing defenses against NSFW content can be easily bypassed, our approach can successfully prevent the generation of NSFW content from both context and adversarial attacks. In summary, our work considers both textual and vision modalities to build a sustainable and effective defense that can be used as a moderator on any open-source or commercial website. And, the proposed work can serve as a valuable baseline for multimodal defense of unsafe content, making the web safe for everyone.

\begin{acks}
    This work was partly supported by Institute for Information \& communication Technology Planning \& evaluation (IITP) grants funded by the Korean government MSIT: (RS-2022-II221199, RS-2024-00337703, RS-2022-II220688, RS-2019-II190421, RS-2023-00230\\337, RS-2024-00356293, RS-2022-II221045, RS-2021-II212068, and RS-2024-00437849). 
\end{acks}



\bibliographystyle{ACM-Reference-Format}
\bibliography{sample-base}

\end{document}